\documentclass[sigconf]{acmart}
\AtBeginDocument{%
  \providecommand\BibTeX{{%
    \normalfont B\kern-0.5em{\scshape i\kern-0.25em b}\kern-0.8em\TeX}}}

\copyrightyear{2025}
\acmYear{2025}
\setcopyright{rightsretained}
\acmConference[KDD '25]{Proceedings of the 31st ACM SIGKDD Conference on
Knowledge Discovery and Data Mining V.1}{August 3--7, 2025}{Toronto, ON,
Canada}
\acmBooktitle{Proceedings of the 31st ACM SIGKDD Conference on Knowledge
Discovery and Data Mining V.1 (KDD '25), August 3--7, 2025, Toronto, ON,
Canada}
\acmDOI{10.1145/3690624.nnnnnnn}
\acmISBN{978-1-4503-XXXX-X/18/06}
\usepackage[ruled, lined, vlined, linesnumbered]{algorithm2e}
\usepackage{amsmath, amsthm}

\usepackage{amssymb}
\usepackage{bm}
\usepackage{multirow}
\usepackage{subfig}
\usepackage{threeparttable}
\usepackage{xcolor, xspace}
\usepackage{algorithmic}
\usepackage{nccmath}
\usepackage{makecell,rotating}
\theoremstyle{definition}

\newcommand{\eg}{\emph{e.g.},\xspace}

\newcommand\figref[1]{Figure~\ref{#1}}
\newcommand\tabref[1]{Table~\ref{#1}}
\newcommand\secref[1]{Sec.~\ref{#1}}

\newcommand\appref[1]{Appendix~\ref{#1}}
\newcommand{\fakeparagraph}[1]{\noindent\textbf{#1.}}

\newcommand{\sysname}{DynAGS\xspace}

\ifodd 1

\newcommand{\xx}[1]{{\color{teal}{#1}}}
\newcommand{\TODO}[1]{\textbf{\color{red}{TODO: #1} }}

\else

\newcommand{\xx}[1]{{#1}
\newcommand{\TODO}[1]{#1}
\fi

\makeatletter
\newcommand{\rmnum}[1]{\romannumeral #1}
\newcommand{\Rmnum}[1]{\expandafter\@slowromancap\romannumeral #1@}
\makeatother

\begin{document}

\title{Dynamic Localisation of Spatial-Temporal Graph Neural Network}

\author{Wenying Duan}
\email{wenyingduan@ncu.edu.cn}
\affiliation{%
  \institution{Jiangxi Provincial Key Laboratory of Intelligent Systems and Human-Machine Interaction, Nanchang University}
  \city{Nanchang}
  \country{China}
}

\author{Shujun Guo}
\email{9109222040@ncu.edu.cn}
\affiliation{%
 \institution{School of Mathematics and Computer Sciences\\Nanchang University}
 \city{Nanchang}
 \country{China}
}

\author{Zimu Zhou}
\affiliation{%
  \institution{School of Data Science\\City University of Hong Kong}
  \city{Hong Kong}
  \country{China}}
\email{zimuzhou@cityu.edu.hk}

\author{Wei Huang}
\email{huangwei@ncu.edu.cn}
\affiliation{%
  \institution{School of Mathematics and Computer Sciences\\Nanchang University}
  \city{Nanchang}
  \country{China}
}

\author{Hong Rao$^*$}
\email{raohong@ncu.edu.cn}
\affiliation{%
 \institution{School of Software\\Nanchang University}
 \city{Nanchang}
 \country{China}
}

\author{Xiaoxi He}\authornote{Corresponding authors: Hong Rao and Xiaoxi He}
\email{hexiaoxi@um.edu.mo}
\affiliation{%
  \institution{Faculty of Science and Technology\\University of Macau}
  \city{Macau}
  \country{China}
}

\begin{abstract}

Spatial-temporal data, fundamental to many intelligent applications, reveals dependencies indicating causal links between present measurements at specific locations and historical data at the same or other locations. 
Within this context, adaptive spatial-temporal graph neural networks (ASTGNNs) have emerged as valuable tools for modelling these dependencies, especially through a data-driven approach rather than pre-defined spatial graphs. 
While this approach offers higher accuracy, it presents increased computational demands. 
Addressing this challenge, this paper delves into the concept of localisation within ASTGNNs, introducing an innovative perspective that spatial dependencies should be dynamically evolving over time. 
We introduce \textit{\sysname}, a localised ASTGNN framework aimed at maximising efficiency and accuracy in distributed deployment. This framework integrates dynamic localisation, time-evolving spatial graphs, and personalised localisation, all orchestrated around the Dynamic Graph Generator, a light-weighted central module leveraging cross attention. 
The central module can integrate historical information in a node-independent manner to enhance the feature representation of nodes at the current moment. This improved feature representation is then used to generate a dynamic sparse graph without the need for costly data exchanges, and it supports personalised localisation.
Performance assessments across two core ASTGNN architectures and nine real-world datasets from various applications reveal that \textit{\sysname} outshines current benchmarks, underscoring that the dynamic modelling of spatial dependencies can drastically improve model expressibility, flexibility, and system efficiency, especially in distributed settings. 

\end{abstract}
%



\maketitle

\section{Introduction}

Spatial-temporal data underpins many contemporary, intelligent web applications. 
Often, these data exhibit spatial and temporal dependencies, indicating that the present measurement at a specific location (in either physical or abstract spaces) is causally dependent on the historical status at the same and other locations. 
Understanding these spatial and temporal dependencies is a crucial aspect of spatial-temporal data mining and is critical to spatial-temporal inference. Consequently, spatial-temporal graph neural networks (STGNNs) have emerged as a potent tool for modelling these dependencies, demonstrating significant success across various fields.

A subclass of STGNN, known as adaptive spatial-temporal graph neural network (ASTGNN), has brought a new perspective to the modelling of these dependencies. Unlike traditional STGNNs that use pre-defined spatial graphs containing prior knowledge, ASTGNNs adopt a data-driven approach. They start with complete graphs and learn the spatial dependency based on the data. This method allows for a more accurate and flexible representation of the spatial-temporal data. However, this approach also introduces higher computational overhead due to the usage of complete graphs.

A recent study~\cite{bib:kdd23:Duan} has proposed a promising solution to this challenge through the concept of \textit{localisation}. Localisation involves pruning the spatial graph to lessen the overall data exchange between nodes and the computational overhead. This reduction is feasible due to the redundancy of information provided by spatial dependency compared to temporal dependency.

\begin{figure*}[t]
    \centering
    \includegraphics[width=0.9\linewidth]{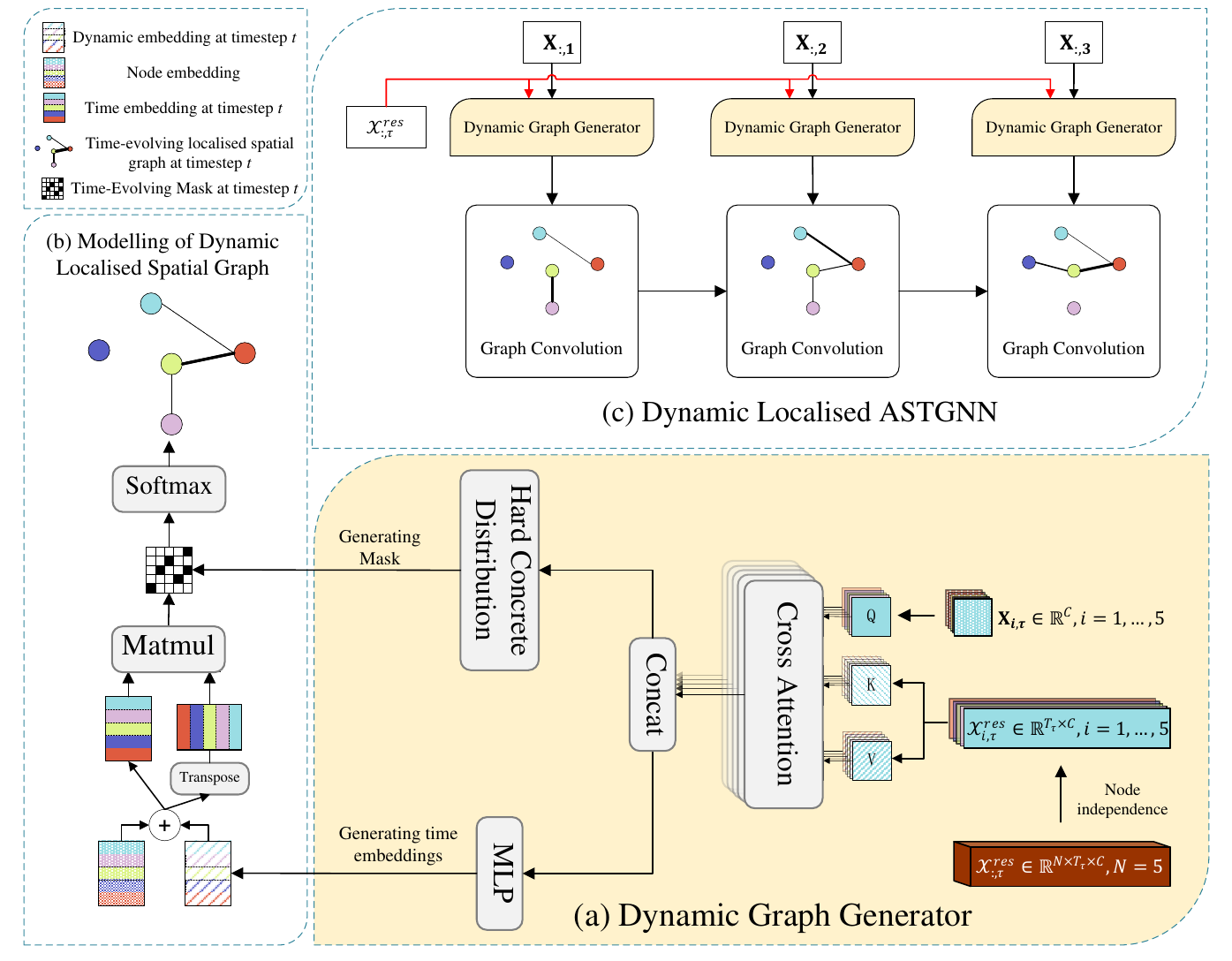}
    \caption{An overview of the \sysname framework with a pre-specified node numbers.}
    \label{fig:overview}
\end{figure*}

In this paper, we explore further the concept of localisation within ASTGNNs. We propose an innovative approach that transcends static spatial dependencies. We argue that the necessary spatial dependencies, modelled by the sparsified spatial graph of an ASTGNN, should not remain static across the entire time interval. Instead, \textbf{these spatial dependencies, and thus the required data exchange between nodes, should be dynamically changing over time}. This concept is illustrated in the section with an orange background in \figref{fig:overview}: the topology and edge weights of the spatial graph are time-evolving.

Moreover, the localisation of ASTGNN is particularly useful for distributed deployments, where a reduction of data exchange between nodes can significantly enhance overall system efficiency. \textbf{Existing dynamic spatial-temporal graph modeling techniques can achieve time-varying sparse graphs, but we argue this will undermine the advantages of localised ASTGNN in distributed deployment. This is because they require node-to-node data exchange at each interval to compute the spatial graph, leading to significant communication overhead, especially with large-scale and high-frequency spatial-temporal data.}
To this end, we proposed \textit{\sysname}, a localised ASTGNN framework designed for optimal efficiency and accuracy in distributed deployment. \sysname has the following features:
\begin{itemize}
    \item \sysname allows for \textbf{dynamic localisation}, which can further decrease the amount of data exchange by a substantial margin. Each node in our framework has the autonomy to decide whether and to which other nodes they need to send data. Importantly, this decision is made using only locally available historical data, making our framework perfectly suited for distributed deployment. 
    \item \sysname introduces a \textbf{time-evolving spatial graph}, in which both the topology of the spatial graph, represented by a mask matrix, and the edge weights of the spatial graph, represented by the adjacency matrix, are dynamic. This feature guarantees optimal expressibility and flexibility of the model, and results in the minimal data exchange necessary to sustain the desired performance. 
    \item \sysname supports \textbf{personalised localisation}, permitting each node to select a personalised trade-off between data traffic and inference accuracy. This flexibility allows each node to maximise its performance based on the available resources, thereby optimising the efficiency and performance of the entire heterogeneous system.
\end{itemize}

To achieve these features, the \textit{\sysname} is designed around a central module known as the Dynamic Graph Generator (DGG). Central to this module is a cross-attention mechanism. This mechanism adeptly amalgamates patch-level historical data with point-level current observational data to synthesize representations pertinent to the present time. Such representations are then utilised to generate time-evolving spatial graphs. These representations adept in the comprehensive integration of multi-scale temporal dependencies. This integration facilitates more nuanced modeling of temporal dependencies, spanning both long-term and short-term spectra, encapsulating elements such as periodicity, proximity, and trend dynamics. Consequently, this leads to an enhanced calibration of incoming weights and connections for each respective node in the system.
An overview of \textit{\sysname} is illustrated in \figref{fig:overview}.
The main contributions of our work are as follows:
\begin{itemize}
    \item We introduce \textit{\sysname}, a novel ASTGNN framework designed for optimal efficiency and inference accuracy in distributed deployment. This framework features dynamic localisation, time-evolving spatial graphs, and personalised localisation to reduce data exchange, enhance model flexibility, and allow nodes to optimise performance based on their own available resources.
    \item The performance of \textit{\sysname} is assessed using two backbone ASTGNN architectures across nine real-world spatial-temporal datasets. 
    The experimental findings indicate that \textit{\sysname} significantly outperforms the state-of-the-art across various localisation degrees, ranging from 80\% to 99.9\%. 
    Specifically, \textit{\sysname}, when operating at a localisation degree of 99.5\%, produces results that are comparable to or even superior to those of the current leading baseline at a localisation degree of 80\%. This results in a considerable decrease in communication overhead, reducing it by no less than 30 times.
    \item  The efficacy of \textit{\sysname} in diverse evaluation experiments substantiates our proposition that spatial dependency is intrinsically time-dependent. This means that the necessary spatial dependencies, and hence the required data exchange between nodes, should continually evolve over time rather than remain static. By embracing this dynamic approach in modelling spatial dependencies within ASTGNNs, we enhance the model's expressibility and flexibility, and optimise system efficiency - a particularly significant improvement in distributed deployments.

\end{itemize}

\section{Related Work}
\subsection{Spatial-Temporal Graph Neural Networks}
STGNNs excel at uncovering hidden patterns in spatial-temporal data \cite{bib:TNNLS20:Wu}, primarily by modeling spatial dependencies among nodes through adjacency matrices. These matrices are constructed using either pre-defined or self-learned methods.

Pre-defined STGNNs use topological structures \cite{guo2021learning, DBLP:conf/ijcai/YuYZ18} or specific metrics (\eg POI similarity) to build graphs \cite{geng2019spatiotemporal, yao2018deep}, but they rely heavily on domain knowledge and graph quality. Given the implicit and complex nature of spatial-temporal relationships, self-learned methods for graph generation have risen in prominence. These methods provide innovative approaches to capture intricate spatial-temporal dependencies, offering a significant advantage over traditional pre-defined models.

Self-learned STGNNs are divided into two primary categories: feature-based and random initialized methods. Feature-based approaches, like PDFormer \cite{DBLP:conf/aaai/JiangHZW23} and DG \cite{DBLP:journals/isci/PengDLLJWZH21}, generate dynamic graphs from time-varying inputs, enhancing model accuracy. Random initialized STGNNs, or ASTGNNs, perform adaptive graph generation via randomly initialized learnable node embeddings.  
Graph WaveNet \cite{bib:IJCAI19:Wu} propose an AGCN layer to learn a normalized adaptive adjacency matrix, and AGCRN \cite{bib:NIPS20:Bai} designs a Node Adaptive Parameter Learning enhanced AGCN (NAPL-AGCN) to learn node-specific patterns. Due to its notable performance, NAPL-AGCN has been integrated into various recent models \cite{DBLP:conf/aaai/Jiang0YJCK0FS23, bib:AAAI22:Choi, bib:ICLR22:Chen}. However, their static node embeddings limit the adaptability of the graphs.
Recent advancements like DGCRN\cite{li2023dynamic} and DMSTGCN \cite{10.1145/3447548.3467275} introduce dynamic node embeddings, allowing for more flexible graph generation in ASTGNNs. Despite these advancements, most self-learned STGNNs generate complete graphs with high computational and communication overheads. Methods such as GTS and ASTGAT mitigate this by employing strategies like Top-K and Gumbel-softmax for partial adjacency matrix retention, enabling discrete and sparse graph generation with dynamic topologies. 

Yet, these approaches still require computation of all node pair connections to determine if there are edges between them, leading to a quadratic increase in data exchange and computational overhead. Our work addresses these challenges in dynamic graph generation, proposing a novel approach in localised STGNNs. By implementing dynamic localisation, we enhance model expressiveness and flexibility, while considerably reducing data exchange requirements and computational overhead. This results in a more efficient balance between model efficiency and accuracy.

\subsection{Graph Sparsification for GNNs}
With the increasing size of graphs, the computational cost of training and inference for GNNs also rises. This escalating cost has spurred interest in graph sparsification, which aims to create a smaller sub-graph while preserving the essential properties of the original graph \cite{bib:kdd23:Duan}. 
SGCN \cite{bib:PAKDD20:Li} is the pioneering work in graph sparsification for GNNs, which involves eliminating edges from the input graph and learning an additional DNN surrogate model. More recent works, such as UGS \cite{bib:ICML21:Chen2} and GBET \cite{bib:AAAI22:You}, examine graph sparsification from the perspective of the lottery ticket hypothesis.

The aforementioned works only studied graph sparsification for standard GNNs with non-temporal data and pre-defined graphs. AGS \cite{bib:kdd23:Duan} extended this concept to spatial-temporal GNNs with adaptive graph architectures, known as the localisation of ASTGNNs. However, AGS localises an ASTGNN by learning a fixed mask, overlooking the fact that spatial dependency varies over time. This approach led to sub-optimal generalisation on unseen data. Unlike AGS, \sysname allows the dynamic localisation of ASTGNNs.
\section{PRELIMINARIES}

\subsection{Notations and Problem Definition}
\label{subsec:notation}
Frequently used notations are summarised in \tabref{tab:notation}.
\begin{table}[h]
\centering
\footnotesize
\caption{The main symbols and definitions in this paper.}
\begin{tabular}{ll}

\Xhline{1pt}
Notation&Definition \\ \hline
$\mathcal{G}$& the spatial graph used for modelling the spatial dependency  \\
$\mathbf{A}$ & the  adjacency matrix of $\mathcal{G}$\\
$N$ & the number of nodes\\
$\mathbf{X}_{i,t}$ & features of the $i$-th node at timestep $t$\\
$\mathbf{A}^{adp}$ &the normalised adaptive adjacency matrix\\
$\mathbf{A}_{t}^{adp}$ &dynamic adaptive adjacency matrix at timestep $t$\\
$\mathbf{M}_{t}$ & dynamic mask to localise ASTGNN at timestep $t$\\
$T_h$ & the look-back period length of the task \\
$d$ & the dimension of node embedding\\
$C$ & the dimension of node feature\\ 
$F$ & the dimension of output feature of ASTGNN\\
$k$ & kernel and stride size of 1D average pooling \\
$\lambda$ &weighting factor of $L_0$-norm\\
\Xhline{1pt}
\end{tabular}
\label{tab:notation}
\end{table}
Following the conventions in related works~\cite{bib:ICONIP18:Seo, bib:IJCAI18:Yu, bib:AAAI18:Yan, bib:IJCAI:bai, bib:IJCAI20:Huang}, we denote the spatial-temporal data as a sequence of frames: 
$\{\mathbf{X}^{1}, \mathbf{X}^{2}, \ldots$  $, \mathbf{X}^{t}, \ldots \}$,
where a single frame $\mathbf{X}^{t}\in \mathbb{R}^{N \times C}$ 
is the $C$-dimensional data collated from $N$ different locations at time $t$.
For a chosen task time $\tau$, we aim to learn a function mapping the ${{T}_{in}}$ historical observations into the future observations in the next ${T}_{out}$ timesteps:
\begin{equation}
    \mathbf{X}^{(\tau+1): (\tau+{T}_{out})}\xleftarrow{}\mathcal{F}(\mathbf{X}^{(\tau-{T}_{in}+1): \tau})
\end{equation}

\subsection{Localised ASTGNN}
\label{subsec:gcn}
\subsubsection{Graph Convolution Network}

A basic method to model the spatial dependency at time $t$ is the Graph Convolutional Network (GCN)~\cite{bib:IJCAI19:Wu,bib:IJCAI:bai,bib:IJCAI20:Huang}.
A graph convolutional layer processing one data frame $\mathbf{X}_{:,t}$ is defined as~\cite{zhang2019graph}:
\begin{equation}
{\mathbf{Z}_{t} =\sigma\left(\tilde{\mathbf{D}}^{-\frac{1}{2}} \tilde{\mathbf{A}} \tilde{\mathbf{D}}^{-\frac{1}{2}}\mathbf{X}_{:,t} \Theta \right)}
\label{eq:gcn}
\end{equation}
where $\sigma(\cdot)$ is the activation function, $\tilde{\mathbf{A}} = \mathbf{I}+\mathbf{A}$ is the adjacency matrix of the graph with added self-connections (with the original adjacency matrix being $\mathbf{A}$), and $\tilde{\mathbf{D}}$ is the diagonal degree matrix of $\tilde{\mathbf{A}}$. 
$ \Theta \in \mathbb{R}^{C \times F}$ is a trainable parameter matrix.
$\mathbf{Z}_{t} \in \mathbb{R}^{N \times F}$ is the layer output, with $F$ indicating the number of output feature dimensions for each node.

\subsubsection{Adaptive Spatial-Temporal Graph Neural Network (ASTGNN)}

A crucial enhancement in modeling the spatial network is the adoption of Adaptive Graph Convolution Networks (AGCNs). These networks capture the dynamics within the graph, paving the way for the development of Adaptive Spatial-Temporal Graph Neural Networks (ASTGNNs) \cite{bib:IJCAI19:Wu,bib:NIPS20:Bai,bib:ICML21:Chen,bib:AAAI22:Choi,bib:ICLR22:Chen}. 

In the subsequent discussion, we provide a brief overview of two representative ASTGNN models: \textit{(i)} The Adaptive Graph Convolutional Recurrent Network (AGCRN) \cite{bib:NIPS20:Bai}; \textit{(ii)} STG-NCDE \cite{bib:AAAI22:Choi}, which is an extension of AGCRN with neural controlled differential equations (NCDEs).

\begin{itemize}

    \item \textbf{AGCRN}. AGCRN enhances the GCN layer by merging the normalised self-adaptive adjacency matrix with Node Adaptive Parameter Learning (NAPL), creating an integrated module known as NAPL-AGCN.
    \begin{equation}
    \begin{aligned}
    \mathbf{A}^{adp}&={softmax}\left({ReLU}\left(\mathbf{E} \cdot \mathbf{E}^\top \right)\right) \\
    \mathbf{Z}_{t}&=\sigma\left(\mathbf{A}^{adp} \mathbf{X}_{:,t} \mathbf{E}\mathbf{W}\right)
    \end{aligned} \label{eq:self-adj}
    \end{equation}
    In this context, $\mathbf{E} \in \mathbb{R}^{N \times d}$, $\mathbf{W} \in \mathbb{R}^{d \times C\times F}$, and consequently $\mathbf{E}\mathbf{W} \in \mathbb{R}^{N \times C\times F}$.
    $\mathbf{A}^{adp} \in \mathbb{R}^{N \times N}$ represents the normalised self-adaptive adjacency matrix \cite{bib:IJCAI19:Wu}.
    $d$ denotes the embedding dimension. 
    Each row of $\mathbf{E}$ embodies the embedding of a node. 
    The embedding of the $i$-th node, represented by the $i$-th row in $\mathbf{E}$, is denoted as $\mathbf{e}_{i}\in \mathbb{R}^{d}$. 
    During training, $\mathbf{E}$ is updated to encapsulate the spatial dependencies among all nodes. 
    Rather than learning the shared parameters $\Theta$ in \eqref{eq:gcn}, NAPL-AGCN employs $\mathbf{E}\mathbf{W}$ to learn parameters specific to each node.
    To capture both spatial and temporal dependencies, AGCRN combines NAPL-AGCN and Gated Recurrent Units (GRU), replacing the MLP layers in NAPL-AGCN with GRU layers.
    
    \item \textbf{STG-NCDE}. STG-NCDE extends AGCRN by incorporating two neural controlled differential equations (NCDEs): a temporal NCDE and a spatial NCDE. 
    
    
\end{itemize}

\subsubsection{Localisation of ASTGNN}
We denote an ASTGNN model as $\mathcal{F}(\cdot;\theta,\mathcal{G})$, where $\theta$ encompasses all the learnable parameters, and $\mathcal{G}$ represents the spatial graph. The graph $\mathcal{G}$ is characterised by its adjacency matrix $\mathbf{A}_{adp}$, as derived from \eqref{eq:self-adj}. The localisation of $\mathcal{F}(\cdot;\theta,\mathcal{G})$ is accomplished by pruning the spatial graph $\mathcal{G}$. This can be mathematically expressed as the Hadamard product of the adjacency matrix $\mathbf{A}_{adp}$ and a mask matrix $\mathbf{M}$. Consequently, the pruned graph has the adjacency matrix $\mathbf{A}_{adp}\odot \mathbf{M}$. During the pruning process, $\mathbf{M}$ is obtained by minimising the following objective:
\begin{equation}\label{loss:2}
\mathcal{L}_{AGS}  =\mathcal{L}(\theta, \mathbf{A}_{adp}\odot \mathbf{M})+
\lambda\left\|\mathbf{M}\right\|_0
\end{equation}
Here, $\mathcal{L}(\theta, \mathbf{A}_{adp}\odot \mathbf{M})$ is the training loss function calculated with the pruned spatial graph, $\left\|\mathbf{M}\right\|_0$ is the $L_{0}$-norm, and the weighting factor $\lambda$ regulates the degree of pruning. As $\left\|\mathbf{M}\right\|_0$ is not differentiable, Adaptive Graph Sparsification (AGS) \cite{bib:kdd23:Duan}, a recent localisation framework for ASTGNNs, resolves this issue by optimising the differentiable approximation of the $L_{0}$-regularisation of $\left\|\mathbf{M}\right\|_0$.

However, in this paper, we contend that a static mask $\mathbf{M}$ and a static adjacency matrix $\mathbf{A}_{adp}$ disregard the fact that spatial dependencies are dynamic over time, resulting in sub-optimal performance on unseen data. Consequently, we aim to achieve dynamic localisation by learning dynamic $\mathbf{M}^{t}\in \mathbb{R}^{N\times N}$ and $\mathbf{A}^{t}_{adp} \in \mathbb{R}^{N\times N}$, which are adapted to the timestep $t$.
\section{Method}

To enable dynamic localisation, it is necessary to generate the time-evolving mask matrix $\mathbf{M}_{t}$ and adaptive adjacency matrix $\mathbf{A}_{t}^{adp}$ as mentioned in \secref{subsec:gcn}. To achieve this, we have designed a framework, \sysname, built around a core module known as the \textit{Dynamic Graph Generator (DGG)}. 
An overview of the \sysname framework is provided in \figref{fig:overview}.

\subsection{Dynamic Graph Generator (DGG)}
\label{sec:nstf}
As outlined in \secref{subsec:notation}, in most spatial-temporal tasks, the look-back period $T_{in}$ is typically much shorter than the entire time interval covered by the available data. Previous studies have predominantly utilized point-level data $\mathbf{X}^{t}$ from the look-back period $T_{in}$  for the generation of dynamic graphs, where $t \in \{\tau-T_{in}+1, \cdots, \tau\}$. This approach, however, lacks the integration of multi-scale information within the temporal dimension and falls short in modeling long-term dependencies.
Accordingly, for the $i$-th node at a chosen task time $\tau$,
our DGG fuses the observations $\mathbf{X}^{t}$ in the look-back period $T_{in}$ with insights from residual historical data by a cross attention.
For the $i$-th node at a chosen task time $\tau$, we define the \textit{residual historical data}:
$\mathcal{X}_{i,\tau}^{res}=\mathbf{X}_{i}^{1:(\tau-T_{in})}\in \mathbb{R}^{T_r\times C}$.

For the $i$-th node at timestep $t \in \{\tau-T_h+1, \cdots, \tau\}$,  a cross attention accepts the node-specific residual historical data $\mathcal{X}_{i,\tau}^{res}$ as key and value, and the observation at the moment $\mathbf{X}_{i,t} \in \mathbb{R}^{C}$ as the query. The cross attention  then output an vector $\mathbf{h}_{i,t} \in \mathbb{R}^{ D}$, which is subsequently utilised to determine the incoming weights and connections of the corresponding node, as detailed in \secref{sec:mainlayer}. Note that in practice, we put an upper-limit on the length $T_r$ of $\mathcal{X}_{i,\tau}^{res}$ based on datasets to ensure that the computational cost is acceptable.
This may change the $\mathcal{X}_{i,\tau}^{res}$ to 
$\mathbf{X}_{i}^{(\tau-T_{in}-T_r+1):(\tau-T_{in})}$.

\subsubsection{Down-sampling and Patching} 

The first challenge lies in efficiently representing the residual historical data, given that the length of this data, $T_r$, is extremely large. Several studies on time-series processing propose reducing the input length via down-sampling or patching. In this case, we employ a combination of both methods to significantly decrease the length of the encoder input. For the $i$-th node, given the residual historical data $\mathcal{X}_{i,\tau}^{res} \in \mathbb{R}^{T_r\times C}$, we first apply 1D average pooling with a kernel $k$ and stride $k$ over $\mathcal{X}_{i,\tau}^{res}$ on the time dimension to down-sample it into $\frac{1}{k}$ of its original slice:

\begin{equation}\label{eq:avgpool}
{\overline{\mathcal{X}}_{i,\tau}^{\frac{{T}_{r}}{k}}=\operatorname{AvgPool1D}\left(\mathcal{X}_{i,t}^{res}\right)} \in \mathbb{R}^{{\frac{T_r}{k}} \times C}
\end{equation}
Next, we divide $\overline{\mathcal{X}}_{i,\tau}^{\frac{{T}_{r}}{k}}$ into $T_{p}$ non-overlapping patches of length $T_s$ to obtain a sequence of the  residual historical patches $\mathbf{X}_{i,\tau}^{p} \in \mathbb{R}^{ T_{p}\times{T_{s} C}}$, $T_{p}=\frac{T_{r}}{kT_{p}}$.
With the use of down-sampling and patching, the number of tokens can be reduced from $T_{r}$ to $\frac{T_{r}}{kT_{p}}$. 

Down-sampling and patching are applied for several reasons: 
\begin{itemize}
    \item To manage residual historical data within time and memory constraints, down-sampling and patching are essential.
    \item Down-sampling effectively retains key periodic and seasonal information from spatial-temporal data, which often includes redundant temporal details \cite{10077946}.
    \item From a temporal perspective, spatial-temporal mining seeks to understand the correlation between data at each time step.
     Patching captures comprehensive semantic information that is not available at the point-level by aggregating time steps into subseries-level patches \cite{DBLP:conf/iclr/NieNSK23}.
\end{itemize}

\subsubsection{Cross Attention}
We utilise a cross-attention to process the residual historical patches and standard historical observations. These patches and observations are initially processed by two trainable linear layers, respectively. During inference, for the $i$-th node at timestep $t \in \{\tau-T_{in}+1, \cdots, \tau\}$:
\begin{equation}
    \mathbf{K} =\mathbf{V}=\mathbf{X}_{i,\tau}^{p}\mathbf{W}_{P}+\mathbf{e}_{pos}, \mathbf{Q}= {\mathbf{X}}_{i}^{t}\mathbf{W}_{q},
\end{equation}
where $\mathbf{W}_{p}\in \mathbb{R}^{T_{s}C\times{d^{\prime}}}$ and $\mathbf{W}_{v}\in \mathbb{R}^{C\times d^{\prime}}$, $\mathbf{e}_{pos}\in \mathbb{R}^{T_{p}\times d^{\prime}}$ is the position embeddings.
Then, we apply cross-attention operations in the temporal dimension to model the interaction between:
\begin{equation}
    \mathbf{h}_{i}^{t} =\operatorname{Softmax}({\frac{\mathbf{Q}\mathbf{K}^{^\top}}{\sqrt{d^{\prime}}}})\mathbf{V} \\
\end{equation}
In the end, the cross-attention outputs $\mathbf{h}_{i,t}$ from the $N$ nodes are stacked together to form a matrix $\mathbf{H}^{t} =\mathbf{h}_{1:N}^{t} $
This matrix $\mathbf{H}^{t}$ is used later to generate the time-evolving $\mathbf{M}^{t}$ and $\mathbf{A}^{t}_{adp}$, which is explained in details in \secref{sec:mainlayer}.

In summary, the attention mechanism within DGG selectively concentrates on sequences of historical data patches most pertinent to the current input. This process culminates in the final fused feature representation, $\mathbf{H}^{t}$, a thorough amalgamation of both current observation and historical data characteristics. 
The influence of historical data on the model's performance is elaborated in \secref{sec:res data}.

\subsection{Dynamic Localised ASTGNNs}\label{sec:mainlayer}

Given a task time $\tau$ (as defined in Section \ref{subsec:notation}), at the timestep $t \in \{\tau-T_{in}+1, \cdots, \tau\}$, we need to facilitate dynamic localisation by generating of the time-evolving mask matrices $\mathbf{M}^{t}$ and adaptive adjacency matrices $\mathbf{A}^{t}_{adp}$.

\subsubsection{Dynamic Mask Generation}

The process of generating the mask matrix, $\mathbf{M}^{t}$, at timestep $t$ begins with a linear transformation applied to $\mathbf{H}^{t}$:
\begin{equation}
{\mathbf{O}_{t}=(\mathbf{H}^{t}\mathbf{W}_{O})}
\end{equation}
where $\mathbf{W}_{O}\in \mathbb{R}^{d^{\prime}\times N}$ is a trainable parameter matrix. The output of this transformation, $\mathbf{O}^{t} \in \mathbb{R}^{N\times N}$, is then utilised to generate $\mathbf{M}^{t}$ through hard concrete distribution sampling. Each entry $o^{t}_{(i,j)} \in \mathbf{O}^{t}$ is converted into binary "gates" $m^{t}_{(i,j)} \in \mathbf{M}^{t}$ using the hard concrete distribution. This distribution is a continuous relaxation of a discrete distribution and can approximate binary values. The computation of the binary "gates" $m^{t}_{(i,j)}$ is expressed as follows:
\begin{equation}
\begin{aligned}
s^{t}_{(i, j)}&= sigmoid\left(\log z-\log (1-z)+\log \left(o^{t}_{(i,j)}\right)\right) / \beta\\
m^{t}_{(i, j)}&=\min \left(1, \max \left(0, s^{t}_{(i, j)}(\zeta-\gamma)+\gamma\right)\right)
\end{aligned}    
\end{equation}
In this equation, $z$ is randomly sampled from a uniform distribution $z \sim \mathcal{U}(0,1)$ and $\beta$ represents the temperature value. Following \cite{bib:kdd23:Duan}, we set $\zeta=-0.1$ and $\gamma=1.1$ in practice.

For the $i$-th node, the corresponding $\mathbf{h}_{i}^{t}$ can be mapped into $\mathbf{m}_{(i,:)}^{t}$, the $i$-th row of $\mathbf{M}^{t}$. This vector $\mathbf{m}_{(i,:)}^{t}$ determines whether the $i$-th node needs to  send data to other nodes, and this decision is solely dependent on $\mathbf{h}^{i}_{t}$. In other words, the $i$-th node can independently decide whether to send data to other nodes without requiring any additional prior data exchange. The mask $\mathbf{M}^{t}$, which determines the topology of the spatial graph, evolves over time as it is computed from the cross attention output $\mathbf{H}^{t}$ at time $t$.

\subsubsection{Dynamic Adaptive Graph Generation}

The NAPL-AGCN employed in ASTGNN only supports a static adjacency matrix, $\mathbf{A}_{adp}$. In order to accommodate a time-evolving $\mathbf{A}_{adp}^t$, we modify the node embedding $\mathbf{E}$ used to generate $\mathbf{A}_{adp}$ in \eqref{eq:self-adj} as follows:
\begin{equation}
    \mathbf{E}^{t} =  \mathbf{E} +\mathbf{\hat{E}}^{t} 
\end{equation}
Here, $\mathbf{\hat{E}}^{t}=\mathbf{H}^{t}\mathbf{W}_{H}$ is the time-evolving embedding, $\mathbf{W}_{H} \in \mathbb{R}^{d^{\prime}\times d}$ are learnable parameters. $\mathbf{\hat{E}}^{t}$ signifies the time-evolving modification applied to the original node embedding $\mathbf{E}$. It's worth noting that even though the computation is expressed in matrix forms, each row's computation in $\mathbf{E}^t$, or $\mathbf{e}^t_{i} \in \mathbb{R}^d$, solely depends on $\mathbf{h}_{i}^{t}$. This implies that the computation of the node-specific time-evolving embedding, $\mathbf{e}^t_{i}$, is performed purely locally without the need for previous data exchange. This process is akin to dynamic mask generation.

To avoid calculating the entire matrix multiplication $\mathbf{E}^t \cdot \mathbf{{E}^t}^\top$ as in \eqref{eq:self-adj}, which would necessitate all nodes to share their own time-evolving embedding $\mathbf{e}^t_{i}$ with each other at every timestep, thereby causing substantial communication overhead, we propose using the locally available time-evolving mask vector $\mathbf{m}_{(i,:)}^{t}$. This vector already determines for the $i$-th node which neighbours it needs to communicate with at time $t$. Consequently, each element $\alpha_t^{(i, j)}$ in the time-evolving adjacency matrix $\mathbf{A}^{t}_{adp}$ is calculated in a similar fashion to the graph attention network (GAT) \cite{DBLP:conf/iclr/VelickovicCCRLB18}:
\begin{equation} \label{eq:adj}
\alpha^t_{(i, j)}=\left\{\begin{aligned}
\frac{\exp \left(\mathbf{e}_{i,t}{\mathbf{e}_{j,t}}^{\top}\right)}{\sum_{k \in \Psi} \exp \left(\mathbf{e}_{i,t}{\mathbf{e}_{k,t}}^{\top}\right)}\quad &, \quad m^{t}_{(i,j)} = 1 \\
0 \qquad \quad \qquad &, \quad m^{t}_{(i,j)}=0
\end{aligned}\right.
\end{equation}
Here $\Psi = \{j\ |\ m^{t}_{(i,j)} = 1\}$ is a set of the $i$-th node's neighbours in the sparse spatial graph. 
Even though the computation in \eqref{eq:adj} still necessitates the exchange of the time-evolving node embedding $\mathbf{e}_{i,t}$ among neighboring nodes, this requirement has been significantly reduced. Moreover, the node embedding $\mathbf{e}_{i,t} \in \mathbb{R}^d$ is quite small in comparison to the volume of data exchanged between adjacent nodes. Consequently, the total communication overhead necessary to support the dynamic spatial graph is minimal.

Finally, we substitute $\mathbf{A}_{adp}^t$ for $\mathbf{A}_{adp}$ in \eqref{eq:self-adj} to obtain the dynamic localised ASTGNN:
\begin{equation}
    \mathbf{Z}^{t}=\sigma\left(\mathbf{A}^{t}_{adp} \mathbf{X}^{t} \mathbf{E}W_{\mathcal{G}}\right)
\end{equation}

\subsubsection{Dynamic Adaptive Graph Sparsification}
Combining the time-evolving mask matrices $\mathbf{M}^{t}$ and adjacency matrices $\mathbf{A}^{t}_{adp}$, we define the procedure for dynamic adaptive graph sparsification as follows:
\begin{equation}\label{loss:3}
\mathcal{L}_{DAGS}=\sum_{t=1}^{T_a}\mathcal{L}_{t}(\theta, \mathbf{A}^{t}_{adp}, \mathbf{M}^t)+
\lambda \sum_{t=1}^{T_a}\left\|\mathbf{M}^{t}\right\|_0
\end{equation}
In this equation, $\mathcal{L}_{t}(\theta, \mathbf{A}^{t}_{adp}, \mathbf{M}^t)$ represents the training loss function computed with the dynamically pruned graph, which is masked by $\mathbf{M}^{t}$ at timestep $t$. $\mathcal{L}_{DAGS}$ is the objective function that can be trained end-to-end using the Adam optimiser.


\subsection{Further Enhancements}
Here we discuss two additional enhancements made to our \sysname, which improve its efficiency further.
\subsubsection{Efficient Dynamic Localisation} 
The generation of a dynamic mask increases the time complexity to $\mathcal{O}(T_{h} \times N^2)$ during inference. To address this, we significantly enhance the efficiency of \sysname by initiating from a sparse spatial graph, pre-localised using AGS. Given a pre-defined graph sparsity $p$, we first train a $p$-localised ASTGNN $\mathcal{F}(\cdot;\theta,\mathbf{A}_{a d p} \odot \mathbf{M}_{p})$ or use a predefined adjacency matrix with a sparsity of  $p$  (if available) to achieve efficient inference. Here, $p$ represents the graph sparsity ($100\%$ means totally sparse), and $\mathbf{M}_{p} \in \mathbb{R}^{N \times N}$ is a mask matrix with a sparsity of $p$. Based on the $p$-localised ASTGNN, we only need to map entries $\{o^{t}_{(i,j)}|o^{t}_{(i,j)} \in \mathbf{O}^{t}, \mathbf{M}_{p}^{(i,j)}\neq 0\}$ into the corresponding $m^{t}_{(i,j)}$. This reduces the time complexity to $\mathcal{O}((1-p) \times T_{h} \times N^2)$ during inference, which is significant considering that $p$ is normally larger than 80\%.

\subsubsection{Personalised Adaptive Graph Sparsification}
In real-world systems, the resources available to different nodes can vary. Consequently, we have implemented personalised localisation instead of global localisation. The node-independent processing of cross attention allows us to assign a unique weighting factor, $\lambda_{i}$, to adjust the degree of localisation for each node $i$:
\begin{equation}
\mathcal{L}_{PDAGS}=\sum_{t=1}^{T_a}\mathcal{L}_{t}(\theta, \mathbf{A}^{t}_{adp}, \mathbf{M}^t)+\sum_{t=1}^{T_a}\sum_{i=1}^{N}\lambda_{i}\left\|\mathbf{m}_{(i,:)}^{t}\right\|_0
\end{equation}
\begin{figure*}[t]   
  \centering            
  \subfloat[AGCRN]
  {
      \label{fig:covid-agcrn}\includegraphics[width=0.9\textwidth]{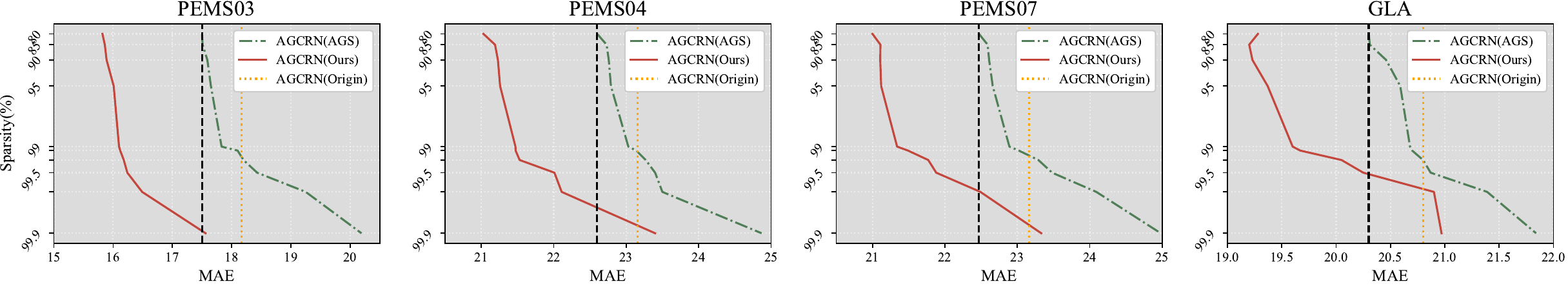}
  }
  
  \subfloat[STG-NCDE]
  {
      \label{fig:covid-stg}\includegraphics[width=0.8\textwidth]{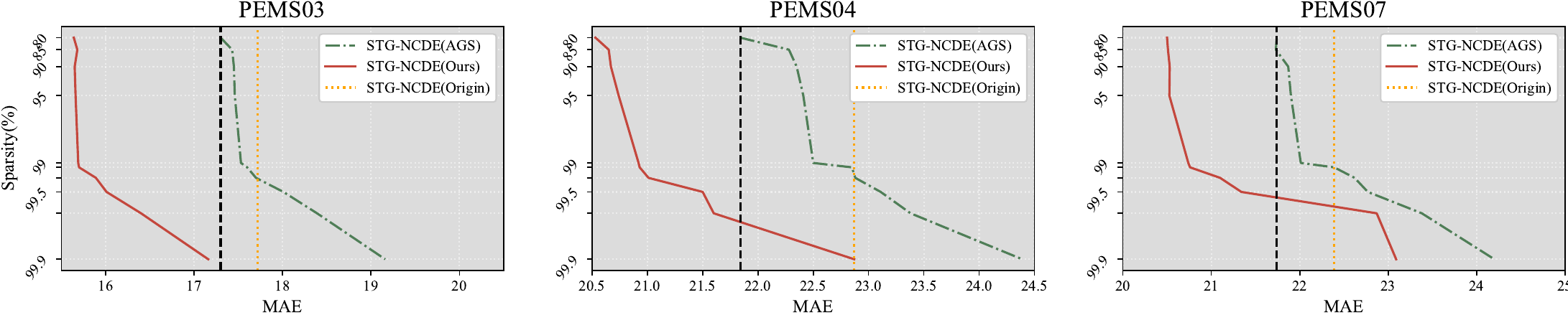}
  }
  \caption{Test accuracies (MAE) of AGCRN (a) and STG-NCDE (b), localised by DynAGS and AGS, evaluated on transportation datasets
 (PEMS03,PEMS04,PEMS07,and GLA).The black vertical dashed lines denote the accuracies of AGCRN
 and STG-NCDE when localised by AGS at a localisation degree of 80\%. RMSE and MAPE results are suppressed for the sake of room and can be found in https://github.com/wenyingduan/DynAGS.}    
  \label{fig:traffic}
\end{figure*}

\subsection{Efficiency Analysis} 
\subsubsection{Complexity Analysis}
Dynamic localization notably decreases the complexity involved in generating dynamic graphs in ASTGNNs.
The inference time complexity of unpruned NAPL-AGCN layers per timestep is $\mathcal{O} (N^2 \times d + L \times N^2 \times F +N \times F^2)$. After sparsification, the inference time complexity of NAPL-AGCN layers is $\mathcal{O} ((1-p) \times N^2 \times d + L \times \left \| \mathbf{M^t} \right \|_0 \times F +N \times F^2)$. where $N$ denotes the number of nodes, $L$ is the number of layers, $d$ is the size of node embedding, $F$ is the size of node feature, and $\left \| \mathbf{M^t} \right \|_0$ is the number of remaining edges.
For more comparative analysis on the complexity of AGS and \textit{\sysname}, please refer to \appref{app:complex}.
\subsubsection{Communication Analysis}
This section presents a theoretical analysis of the communication costs for \sysname and unpruned ASTGNNs.
For unpruned ASTGNNs, the communication overhead during inference is:
\begin{equation}
    {\underbrace{L\times N^2 \times d}_{\text{graph generation terms}}} +  \underbrace{L\times N^2 \times F}_{\text{graph convolution terms}},
\end{equation}
In contrast, \textit{\sysname} significantly reduces this overhead:
\begin{equation}
    {\underbrace{L\times\left \| \mathbf{M^t} \right \|_0  \times d}_{\text{graph convolution terms}}} +  \underbrace{L\times  \left \| \mathbf{M^t} \right \|_0 \times F}_{\text{graph generation terms}}.
\end{equation}
We also experiment the reduction of communication costs in real-world application in \secref{sec:eff}.

\section{Experiments}
\begin{figure*}[t]   
  \centering            
  \subfloat[AGCRN]
  {
      \label{fig:covid-agcrn}
      \includegraphics[width=0.45\textwidth]{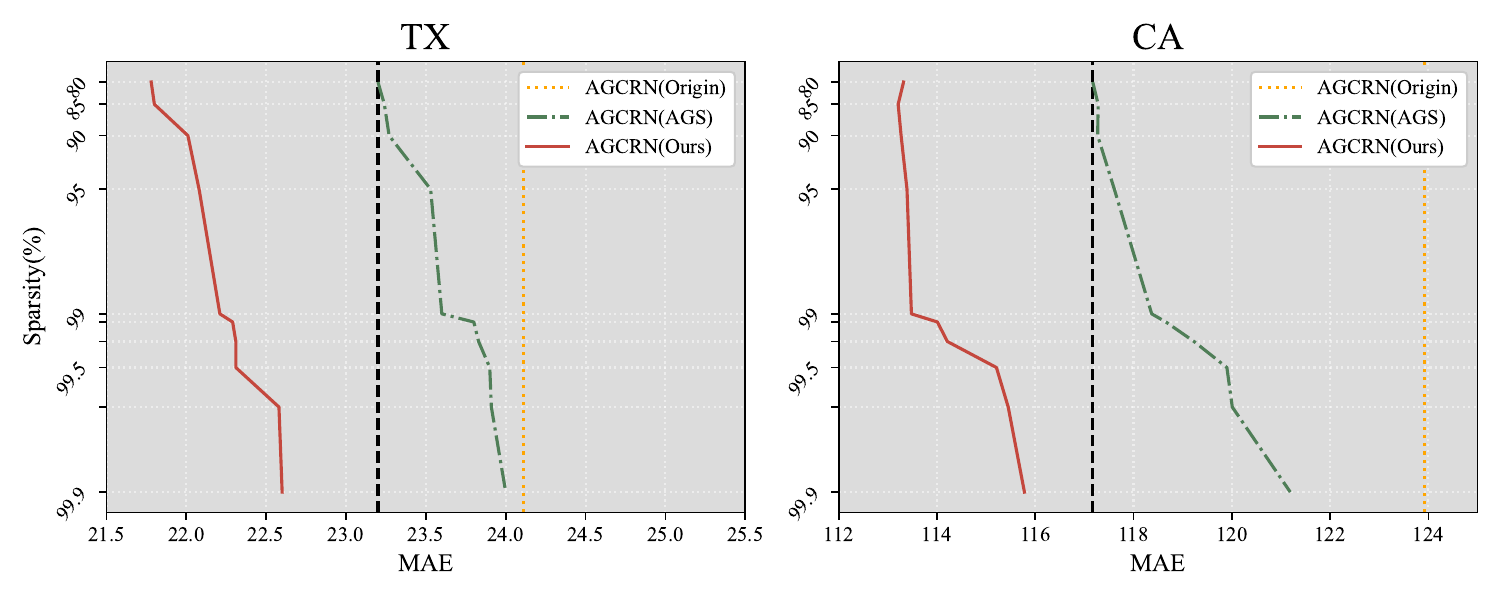}
  }
  \subfloat[STG-NCDE]
  {
      \label{fig:covid-stg}
      \includegraphics[width=0.45\textwidth]{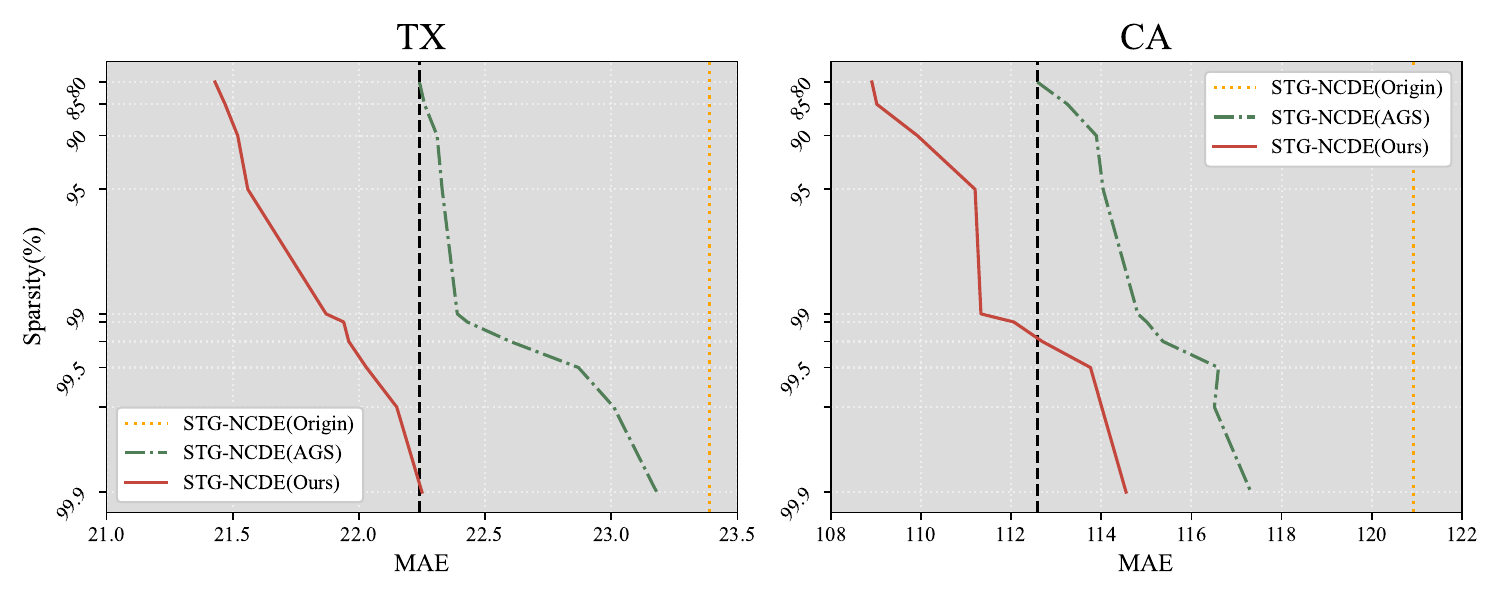}
  }
   \caption{Test accuracies (MAE) of AGCRN (a) and STG-NCDE (b), localised by DynAGS and AGS, evaluated on biosurveillance datasets.The black vertical dashed lines denote the accuracies of AGCRN
 and STG-NCDE when localised by AGS at a localisation degree of 80\%.}    
  \label{fig:covid}   
\end{figure*}

\subsection{Experimental Setup}
\fakeparagraph{Datasets} \sysname was evaluated on nine real-world spatial-temporal datasets from four web applications: transportation, blockchain, and biosurveillance forecasting.
For a more comprehensive evaluation of the generalisation ability, particularly extrapolation to new data, it was necessary to introduce distribution shifts into the training and testing data. This goal was achieved by using more challenging settings rather than standard ones in these domains:
\begin{itemize}
    \item The training/validation/testing sets are split by time. Consequently, less data is used for both training and testing. This approach creates a larger time gap between the training set and testing set than the standard setting. This type of dataset splitting introduces a distribution shift between training and testing data, as the distributions of spatial-temporal data change over time.
    \item We use longer input and output lengths for historical observations (inputs) to predict future observations (outputs). The spatial-temporal dependencies of inputs and outputs may differ over a large time span, as time progresses. This approach can introduce a more significant temporal shift between input and output.
\end{itemize}

Detailed information for each datasets and their configurations are provided in \appref{app:datasets}.

\fakeparagraph{Baseline} 
AGS\cite{bib:kdd23:Duan} was selected as the sole baseline because, to the best of our knowledge, AGS is currently the only method specifically designed for the localisation of ASTGNNs. Moreover, it represents a state-of-the-art method that was published recently. We benchmarked the performance of \sysname against AGS using two representative ASTGNN backbone architectures: 
\textbf{AGCRN} and \textbf{STG-NCDE}. AGCRN stands out as the most representative ASTGNN for spatial-temporal forecasting, while STG-NCDE serves as a state-of-the-art extension of AGCRN.

\fakeparagraph{Implementation} 
The dataset-specific hyperparameters of AGCRN and STG-NCDE, followed the same setup as in the original papers. Other dataset-specific hyperparameters for reproducibility are provided in \appref{app:detail}.

\begin{figure*}[htbp]
    \centering
    \includegraphics[width=0.85\linewidth]{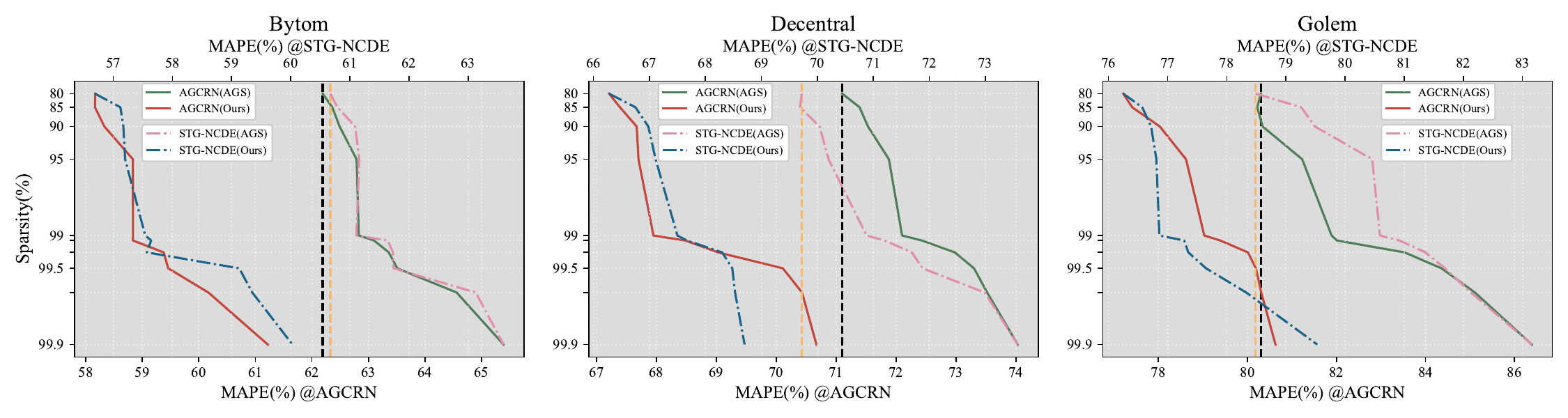}
    \caption{Test accuracies (MAPE) of AGCRN and STG-NCDE, localised by \textit{\sysname} and AGS, evaluated on blockchain datasets. The black and yellow vertical dashed lines denote the MAPE of AGCRN and STG-NCDE when localised by AGS at a localisation degree of 80\%, respectively. The MAPE of the original models are omitted, as the localised ones are always better.}
    \label{fig:eth}
\end{figure*}
\subsection{Test Accuracies}\label{sec:results}
Test accuracies (MAE) of AGCRN and STG-NCDE, localized by \textit{\sysname} and AGS across localisation degrees from $80\%$ to $99.9\%$, are presented in \figref{fig:traffic} , \figref{fig:covid} and \figref{fig:eth} for transportation,biosurveillance and blockchain datasets, respectively.
From these results, the following observations can be made:
\begin{itemize}
    \item \textit{\sysname} is adept at capturing dynamic spatial-temporal dependencies and delivers promising predictions for unseen data. Additionally, our \textit{\sysname} enhances the performance of AGS noticeably. For the same localisation degree, ours surpasses AGS by up to $13.5\%$ less error across all datasets.
    \item \textit{\sysname} excels in the trade-off between data exchange and accuracy. When compared to AGS at a localisation degree of $80\%$, \textit{\sysname} at a localisation degree of $99.5\%$ frequently attains comparable or even superior accuracies, implying that nearly 40 times less communication is needed to sustain the same performance.
    \item 
    The results from the GLA dataset indicate that \textit{\sysname} consistently outperforms AGS. This demonstrates that \textit{\sysname} can be scaled up to large-scale spatial-temporal data.
    \item The accuracy of \textit{\sysname} does not always decrease as sparsity increases (\eg \figref{fig:traffic-agcrn}). We believe this is due to the regularization effect of sparsifying the spatial graph, which suggests that non-localized AGCRNs and STG-NCDEs suffer from overfitting to a certain degree.

\end{itemize}

\subsection{Impact on Resource Efficiency}\label{sec:eff}
As highlighted in \secref{sec:results}, an ASTGNN localised via \textit{\sysname} to the degree of 99.5\% (average degree over $T_{in}$ timesteps) delivers results comparable or even better than that with AGS to a localisation degree of 80\%. Thus, we evaluated the computational demand during the inference of 99.5\%-localised AGCRN and STG-NCDE via \textit{\sysname} and during the inference of 80\%/99.5\%-localised AGCRN and STG-NCDE via AGS. \textbf{We also simulated the distributed computation of \textit{\sysname} using Python's multiprocessing library and compared the communication cost during inference of 99.5\%-localised AGCRN via \textit{\sysname} and during the inference of 80\%/99.5\%-localised AGCRN via AGS, as shown in \tabref{tab:comm}.} The detailed description of simulation can be found in \appref{app:sim}.

Our findings reveal that \textit{(i)} both \textit{\sysname} and AGS can effectively reduce the computation required for inference, \textit{(ii)} comparing \textit{\sysname} and AGS shows that \sysname has a marginally higher computing cost for AGCRN but is more efficient for STG-NCDE when performance is similar, and \textit{(iii)} \textit{\sysname} cuts communication cost by a factor of nearly 40 relative to AGS. Additionally, when the sparsity is the same, the computational overhead of \textit{\sysname} is still slightly higher than that of AGS, but the performance of the former is significantly better than the latter.

In real-world distributed deployment scenarios, a significant portion of resources is allocated to communication between nodes, often overshadowing local computation costs \cite{DBLP:conf/icde/LiDK09, bib:mb:li2019}. In this light, the marked reduction in communication cost by \sysname becomes especially significant. The minor rise in local computation by \sysname is less concerning, given that local computation is generally less resource-intensive than inter-node communication.

Thus, these outcomes affirm that the advantages of \sysname, predominantly the substantial communication cost savings, more than compensate for the occasional minor upticks in computational demand. This positions \sysname as a more resource-efficient solution in real-world distributed deployment contexts.




\section{Ablation Studies}

\begin{table*}[htbp]
\footnotesize
\caption{Computation cost during inference. The localisation degree also indicates the reduction ratio of communication cost.}
\label{tab:cost}
\begin{tabular}{lcccclcc}
\Xhline{1pt}
Methods                      & \multicolumn{7}{c}{Computation Cost for Inference (GFLOPs)}                                 \\ \cline{2-8} 
                             & PEMS03      & PEMS04      & PEMS07      &GLA    &By/De/Go       & CA         & TX         \\ \hline
Original AGCRN               & 5.40        & 4.25        & 24.71       & 192.69     & 0.16       & 0.89       & 5.49       \\
80\%-localised AGCRN(AGS)    & 2.26($\times$2.40\textcolor{green}{$\uparrow$})  & 1.93($\times$2.20\textcolor{green}{$\uparrow$})  & 5.57($\times$4.43\textcolor{green}{$\uparrow$})  & 48.18($\times$4.00\textcolor{green}{$\uparrow$}) & 0.09($\times$1.78\textcolor{green}{$\uparrow$}) & 0.81($\times$1.11\textcolor{green}{$\uparrow$}) & 3.69($\times$1.49\textcolor{green}{$\uparrow$}) \\
99.5\%-localised AGCRN(AGS) 
&2.05($\times$2.63\textcolor{green}{$\uparrow$}) 
&1.80($\times$2.36\textcolor{green}{$\uparrow$}) &4.07($\times$6.07\textcolor{green} {$\uparrow$}) 
&12.96($\times$14.86\textcolor{green} {$\uparrow$})
&0.09($\times$1.78\textcolor{green}{$\uparrow$})
&0.78($\times$1.14\textcolor{green} {$\uparrow$})
&3.64($\times$1.51\textcolor{green} {$\uparrow$})\\
99.5\%-localised AGCRN(Ours)  
&2.63($\times$2.05\textcolor{green}{$\uparrow$})  
&2.26($\times$1.88\textcolor{green}{$\uparrow$}) 
&6.51($\times$3.80\textcolor{green}{$\uparrow$})
& 13.52($\times$14.24\textcolor{green}{$\uparrow$})
&0.12($\times$1.39\textcolor{green}{$\uparrow$})
&0.85($\times$1.05\textcolor{green}{$\uparrow$}) & 3.88($\times$1.41\textcolor{green}{$\uparrow$}) \\ \hline
Original STG-NCDE            & 21.63       & 18.55       & 53.36       &--      & 3.16       & 0.47       & 2.14       \\
80\%-localised STG-NCDE(AGS) & 15.96($\times$1.36\textcolor{green}{$\uparrow$}) & 14.38($\times$1.28\textcolor{green}{$\uparrow$}) & 18.87($\times$2.83\textcolor{green}{$\uparrow$}) & -- & 2.91($\times$1.09\textcolor{green}{$\uparrow$}) & 0.44($\times$1.08\textcolor{green}{$\uparrow$}) & 1.44($\times$1.48\textcolor{green}{$\uparrow$}) \\
99.5\%-localised STG-NCDE(AGS) &14.58($\times$1.48\textcolor{green}{$\uparrow$}) &13.37($\times$1.39\textcolor{green}{$\uparrow$}) &10.46($\times$5.10\textcolor{green}{$\uparrow$}) & --&2.90($\times$1.09\textcolor{green}{$\uparrow$}) &0.44($\times$1.08\textcolor{green}{$\uparrow$})&1.36($\times$1.57\textcolor{green}{$\uparrow$})\\
99.5\%-localised STG-NCDE(Ours)        & 14.93($\times$1.45\textcolor{green}{$\uparrow$}) & 13.67($\times$1.35\textcolor{green}{$\uparrow$}) & 11.2($\times$4.76\textcolor{green}{$\uparrow$})  & 8.85($\times$1.16\textcolor{green}{$\uparrow$}) & 2.96($\times$1.07\textcolor{green}{$\uparrow$}) & 0.46($\times$1.02\textcolor{green}{$\uparrow$}) & 1.42($\times$1.50\textcolor{green}{$\uparrow$}) \\ \Xhline{1pt}
\end{tabular}
\end{table*}

\begin{table*}[htbp]
\caption{Performance of 99.5\%-localised AGCRNs via DynAGS compared with other non-localised ASTGNN architectures.}
\begin{tabular}{llllllllll}
\Xhline{1pt}
Model                            & \multicolumn{3}{l}{PEMS03} & \multicolumn{3}{l}{PEMS04} & \multicolumn{3}{l}{PEMS07} \\ \hline
Metric                           & MAE    & RMSE   & MAPE(\%) & MAE    & RMSE   & MAPE(\%) & MAE    & RMSE   & MAPE(\%) \\ \hline
Z-GCNET                          & 18.59  & 34.08  & 18.72    & 23.15  & 36.21  & 15.97    & 26.47  & 42.13  & 11.23    \\
STG-NCDE                         & 17.83  & 33.72  & 17.51    & 23.44  & 35.34  & 15.21    & 25.87  & 40.41  & 10.63    \\
TAMP-S2GCNets                    & 17.72  & 32.74  & 16.08    & 23.40  & 36.43  & 14.88    & 22.39  & \textbf{37.63}  & 9.82     \\
99.5-localised AGCRN(via DynAGS) & \textbf{16.24}  & \textbf{32.61}  & \textbf{15.59}    & \textbf{22.87}  & \textbf{34.84}  & \textbf{15.53}    & \textbf{21.88}  & 38.01  & \textbf{9.53}     \\ \Xhline{1pt}
\end{tabular}\label{tab:vs. non-localised}
\end{table*}

\begin{table}[htbp]
\footnotesize
\caption{Communication cost during inference (GB).}
\begin{tabular}{llll}
\Xhline{1pt}
Method & AGCRN  & 80\%-localised(AGS)  & 99.5\%-localised(AGS \&Ours) \\ \hline
GLA    & 168.22 & 33.64 & 0.91   \\
PEMS07 & 8.92   & 1.78  & 0.05   \\ 
\Xhline{1pt}
\end{tabular}\label{tab:comm}
\end{table}

\subsection{Impact of Residual Historical Data and Dynamic Graph}\label{sec:res data}
We evaluated the effects of residual historical data and dynamic graph on AGCRN (on PEMS07, Bytom, and TX datasets) by comparing \sysname with its variants, \sysname$\dag$ and \sysname$\ddag$. The variant \sysname$\dag$ is obtained by removing the cross attention, thereby eliminating the residual historical data. In contrast, the \sysname$\ddag$ variant uses only the node embedding $\mathbf{E}$ to generate a static mask, similar to AGS, resulting in a fixed spatial graph topology over time.

As observed from \tabref{ablation:pems}, \sysname consistently outperforms \sysname$\dag$ and \sysname$\ddag$ by a significant margin. These results confirm that  \textit{(i)} the large-scale temporal features and global information provided by the residual historical data contribute positively to the performance of \sysname, and \textit{(ii)} dynamic graph modeling, which captures evolving spatial-temporal dependencies, significantly improves prediction performance.


\subsection{DynAGS vs. Other Non-Localised ASTGNNs}
Recent advancements in ASTGNNs have introduced improved architectures of AGCRN, including Z-GCNETs \cite{bib:ICML21:Chen}, STG-NCDE \cite{bib:AAAI22:Choi}, and TAMP-S2GCNets \cite{bib:ICLR22:Chen}, tailored for various applications.
This motivates us to evaluate how our localised AGCRNs compare to these advanced variations.Results are shown in \tabref{tab:vs. non-localised}.
Our results demonstrate that the localised AGCRNs consistently achieve superior inference performance, even surpassing state-of-the-art architectures, while significantly reducing computational complexity, thereby underscoring their efficacy and practicality.

\begin{table}[t]
\footnotesize
\caption{Impact of residual historical data and dynamic graph on the performance of the PEMS07, Bytom, and TX datasets.}
\label{ablation:pems}
\begin{tabular}{@{\extracolsep{1pt}}lllllll@{}}
\Xhline{1pt}
\multirow{2}{*}{Method}                 & \multirow{2}{*}{Sparsity}      & \multicolumn{2}{c}{PEMS07}      &Bytom             & \multicolumn{2}{c}{TX}          \\ \cline{3-4} \cline{5-5} \cline{6-7} 
                                        &                               & MAE            & MAPE(\%)       & MAPE(\%)       & MAE            & MAPE(\%)       \\ \hline
\multirow{3}{*}{\makecell{AGCRN \\(\sysname)}}        
                                        & 50\%   & \textbf{21.63} & \textbf{9.22} & \textbf{62.36} & \textbf{23.24} & \textbf{96.21} \\
                                        & 80\%                      & \textbf{21.00} & \textbf{8.91} & \textbf{61.99} & \textbf{23.20} & \textbf{96.80} \\
                                        & 99\%     & \textbf{21.29} & \textbf{10.43} & \textbf{65.10} & \textbf{23.80} & \textbf{98.30} \\ \hline
\multirow{3}{*}{\makecell{AGCRN\\(\sysname$\dag$)}}
                                        & 50\%        & 23.63          & 11.03          & 65.12          & 24.50          & 98.21          \\
                                        & 80\%                  & 22.89          & 10.57          & 65.23          & 24.37          & 98.42          \\
                                        & 99\%        & 23.17          & 10.86          & 66.78          & 24.77          & 101.69         \\ \Xhline{1pt}
\multirow{3}{*}{\makecell{AGCRN\\(\sysname$\ddag$)}}
& 50\%& 24.06& 11.41& 66.07& 24.99& 100.09\\
& 80\%& 23.17& 11.27& 67.63& 24.24& 99.53\\
& 99\%& 24.29& 11.95& 67.98& 26.01 & 103.79\\ 
\Xhline{1pt}
\end{tabular}
\end{table}

\subsection{Personalised Localisation}

To further validate our personalised localisation approach, we conducted an additional ablation study. We divided all nodes into distinct groups, with each group assigned a specific in-degree compression ratio. Instead of merely testing these groups, we introduced
a variation: while keeping a group’s in-degree compression ratio unchanged, we adjusted the ratios of the other groups. The results from this exercise were clear. The performance of a particular group remained unaffected by the in-degree sparsity alterations of
other groups.

This experimental evidence underscores the fact that
personalising the localisation of ASTGNNs doesn’t detract from performance, but rather presents a compelling and practical method.
Further details of this experiment can be found in \appref{app: personal}.

\section{Conclusion}

This study introduced \textit{\sysname}, an innovative ASTGNN framework that dynamically models spatial-temporal dependencies. Through the integration of dynamic localisation and a time-evolving spatial graph, \textit{\sysname} has proven superior in performance across various real-world datasets, demonstrating significant reductions in communication overhead. Our findings underscore the intrinsic time-dependent nature of spatial dependencies and advocate for a shift from static to dynamic representations in spatial-temporal data analysis. Looking ahead, \textit{\sysname} paves the way for future advancements in the realm of spatial-temporal data mining, emphasising both adaptability and efficiency.
\section{Acknowledgment}
We thank the reviewers for their constructive comments. 
Wenying Duan and Shujun Guo's research was supported in part
by Jiangxi Provincial Natural Science Foundation No. 20242BAB20066.
Wei Huang's research was supported by NSFC No.62271239, and Jiangxi Double Thousand Plan No.JXSQ2023201022.
Zimu Zhou's research was supported by Chow Sang Sang Group Research Fund No. 9229139.  Sincere thanks to Xiaoxi and Wenying’s mutual friend, Te sha, for providing the academic exchange platform.

\bibliographystyle{ACM-Reference-Format}
\clearpage
\bibliography{cites}
\appendix
\section{Appendix}

\subsection{Additional Dataset Details}\label{app:datasets}

\begin{table}[t]
\centering
\caption{Summary of datasets used in spatial-temporal forecasting.}\label{tab:datasets}
\begin{tabular}{lcc}
    \toprule
     Datasets&\#Nodes& Range\\
    \midrule
     PEMS03&358&09/01/2018 - 30/11/2018\\ 
     PEMS04&307&01/01/2018 - 28/02/2018\\
     PEMS07&883&01/07/2017 - 31/08/2017\\
     GLA&3,834&01/01/2019 - 31/12/2019\\
     \midrule
     Bytom&100&27/07/2017 - 07/05/2018\\
     Decentral&100&14/10/2017 - 07/05/2018\\
     Golem&100&18/02/2017 - 07/05/2018\\
     \midrule
     CA&55&01/02/2020 - 31/12/2020\\
     TX&251&01/02/2020 - 31/12/2020\\
     \bottomrule
\end{tabular}
\end{table}
 
   

\tabref{tab:datasets} summarise the specifications of the datasets used in our experiments.
The detailed datasets and configurations are as follows:
\begin{itemize}
    \item \textbf{Transportation}: We analyze three widely-studied traffic forecasting datasets from the Caltrans Performance Measurement System (PEMS): \textbf{PEMS03}, \textbf{PEMS04}, and \textbf{PEMS07} \cite{bib:others01:Chen}. Additionally, we use the \textbf{GLA} (Greater Los Angeles) dataset from the recent LargeST benchmark \cite{liu2023largest}, currently recognized as the second largest spatio-temporal dataset. In our experiments, PEMS03, PEMS04, PEMS07, and GLA are partitioned in the ratio of 4.5:4:1.5 for training, validation, and testing, respectively. This partitioning scheme deviates from the standard ratio of 6:2:2. For PEMS03, PEMS04, and PEMS07, traffic flows are aggregated in 5-minute intervals, and we perform a 24-sequence-to-24-sequence forecasting, differing from the standard 12-sequence-to-12-sequence approach. 
    For GLA, traffic flows are aggregated in 15-minute intervals. Here, we adhere to the standard 12-sequence-to-12-sequence forecasting format for this dataset. 
    Model accuracy is assessed using the Mean Absolute Error (MAE), Root Mean Square Error (RMSE), and Mean Absolute Percentage Error (MAPE).

    \item \textbf{Blockchain}: We use three Ethereum price datasets: \textbf{Bytom}, \textbf{Decentral}, and \textbf{Golem}~\cite{bib:SMD20:Li}. These datasets are represented as graphs, where nodes and edges signify user addresses and digital transactions, respectively. The interval between two consecutive timestamps is one day. In our settings, Bytom, Decentraland, and Golem token networks are divided with a ratio of 6:3:1 for training, validation, and testing. 
    This differs from the original setting, which uses a ratio of 8:2 for training and testing \cite{bib:ICLR22:Chen}.
    We use 14 days of historical data to forecast the subsequent 14 days. This differs from the original setting, which uses 7 days historical data to predict future 7 days data.  Given that MAE and RMSE values for these datasets are exceedingly small and don't reflect performance accurately, only MAPE is used, following \cite{bib:ICLR22:Chen}.
    
    \item \textbf{Biosurveillance}: We adopt the California (\textbf{CA}) and Texas (\textbf{TX}) COVID-19 biosurveillance datasets \cite{bib:ICLR22:Chen} for forecasting the number of hospitalised patients. The data's time interval is one day. CA and TX datasets are split with a 6:3:1 ratio for training, validation, and testing. 6 days of historical data are used to predict the next 30 days. These differ from the original settings \cite{bib:ICLR22:Chen}, which split CA and TX datasets with an 8:2 ratio for training and testing and use 3 days of historical datat to predict the next 15 days. Both MAE and MAPE are utilised as accuracy metrics.
    
\end{itemize}

\subsection{Comparative Analysis on the Complexity of \sysname and AGS}\label{app:complex}

The time complexity of \sysname is:
\begin{equation}
\begin{split}
    {\underbrace{T_s \times L \times C^2}_{\text{cross attention terms}}} +  \underbrace{K \times L \times \mathcal{O} ((1-p) \times N^2 \times d}_{\text{graph generation terms}} \\+ \underbrace{K \times L \times \left \| \mathbf{M^t} \right \|_0 \times F +N \times F^2}_{\text{pruned graph convolution terms}},
\end{split}
\end{equation}
while the time complexity of AGS is $K \times L \times \left \| \mathbf{M^t} \right \|_0 \times F +N \times F^2$. In contrast, the time complexity of unpruned ASTGNN is $K \times L \times N^2 \times F +N \times F^2$.

It is evident that the additional time complexity introduced by \sysname is $T_s \times L \times C^2 + K \times L \times \mathcal{O}((1-p) \times N^2 \times d)$,  which encompasses the cross attention terms and graph generation terms.  Given that $N \gg T_s$ and $N \gg L$, we have $T_s \times L \times C^2 + K \times L \times \mathcal{O}((1-p) \times N^2 \times d) \ll K \times L \times N^2 \times F + N \times F^2$. 
As mentioned in \secref{sec:results}, compared to AGS at a localization degree of 80\%, \textit{\sysname} at a localization degree of 99.5\% achieves comparable or even superior accuracies. This demonstrates that \textit{\sysname} can effectively reduce the time complexity of ASTGNN, as the time complexity is reduced by $99.5\% \times K \times L \times N^2 \times F$, given that the computational overhead of graph convolution is significantly higher than the additional complexity introduced by \textit{\sysname}. Furthermore, there is no data exchange during the computation of cross attention and graph generation. Overall, \textit{\sysname} provides a highly efficient sparse dynamic graph learning method that requires very few resources.

\subsection{Experiment Details}\label{app:detail}
The hyperparameter setups specific to each dataset are provided in \tabref{tab:hyperparam}. We optimise all models across all datasets using the Adam optimiser for up to 200 epochs. We employ an early stopping strategy with a patience of 30 epochs.

\begin{figure}[htbp]
   \centering
    \includegraphics[width=1\linewidth]{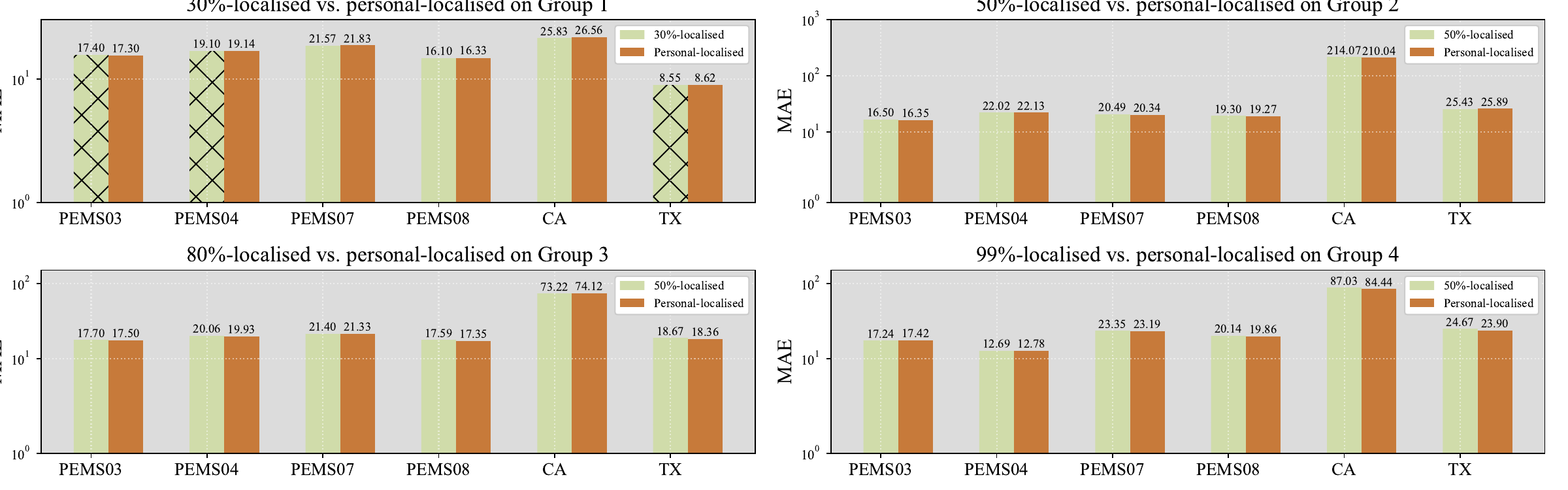}
    \caption{Results of personalised localisation.}
    \label{fig:personal}
\end{figure}

\begin{table*}[b]

\caption{Dataset-specific hyperparameter setups.}
\label{tab:hyperparam}
\begin{tabular}{llllllccc}
\Xhline{1pt}
\multirow{2}{*}{Dataset Type} & \multirow{2}{*}{Datasets} & \multirow{2}{*}{\makecell{Max length of \\rest historic data}} & \multirow{2}{*}{\makecell{Down-sample\\kernel size}} & \multirow{2}{*}{\makecell{Patch\\length}} & \multirow{2}{*}{\makecell{Batch\\size}} & \multicolumn{3}{c}{Learning rate} \\ \cline{7-9} 
                           &                           &                                                &                                          &                               &                             & AGCRN      & STG          \\ \hline
\multirow{4}{*}{Traffic}   & PEMS03                    & 4032                                           & 12                                       & 24                            & 64                          & 1e-3       & 1e-3          \\
                           & PEMS04                    & 4032                                           & 12                                       & 24                            & 64                          & 1e-3       & 1e-3          \\
                           & PEMS07                    & 4032                                           & 12                                       & 24                            & 64                          & 1e-3       & 1e-3         \\
                           & GLA                    & 4032                                           & 4                                       & 24                            & 32                          & 1e-3       & --         \\ \hline
\multirow{3}{*}{Ethereum}  & Decentraland              & 56                                             & 1                                        & 7                             & 8                           & 1e-2       & 1e-2         \\
                           & Bytom                     & 56                                             & 1                                        & 7                             & 8                           & 1e-2       & 1e-2          \\
                           & Golem                     & 56                                             & 1                                        & 7                             & 8                           & 1e-2       & 1e-2           \\ \hline
\multirow{2}{*}{COVID-19}  & TX                        & 56                                             & 1                                        & 7                             & 16                          & 1e-1       & 1e-2         \\
                           & CA                        & 56                                             & 1                                        & 7                             & 16                          & 1e-2       & 1e-2          \\  \Xhline{1pt}
\end{tabular}
\end{table*}

\subsection{Computing Infrastructure}
All experiments are implemented using Python and PyTorch 1.8.2. Training and testing were executed on a server equipped with an Nvidia A6000 (48GB memory). The simulation of distributed computing was executed on a server equipped with 96 Intel 6348H 2.3GHz CPU cores and 1024GB of memory.

\subsection{Additional Experimental Results}
\begin{figure*}[ht]
    \centering
   \includegraphics[width=0.9\linewidth]{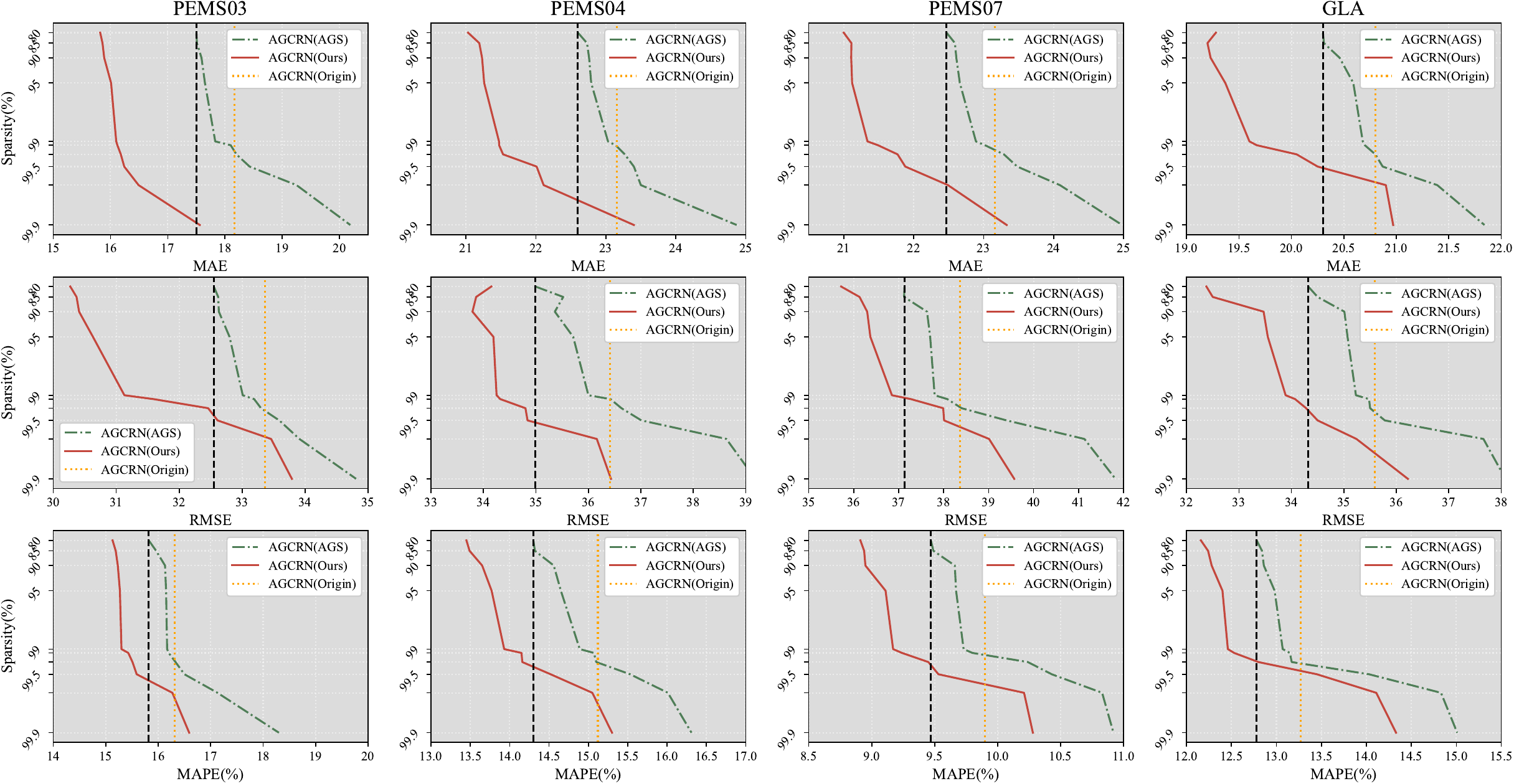}
    \caption{Test accuracies of AGCRN, localised by \textit{\sysname} and AGS, evaluated on transportation datasets (PEMS03, PEMS04, PEMS07, and GLA). The black vertical dashed lines denote the accuracies of AGCRN when localised by AGS at a localisation degree of 80\%.
    }
    \label{fig:traffic-agcrn}
\end{figure*}

\begin{figure*}[htbp]
    \centering
    \includegraphics[width=0.8\linewidth]{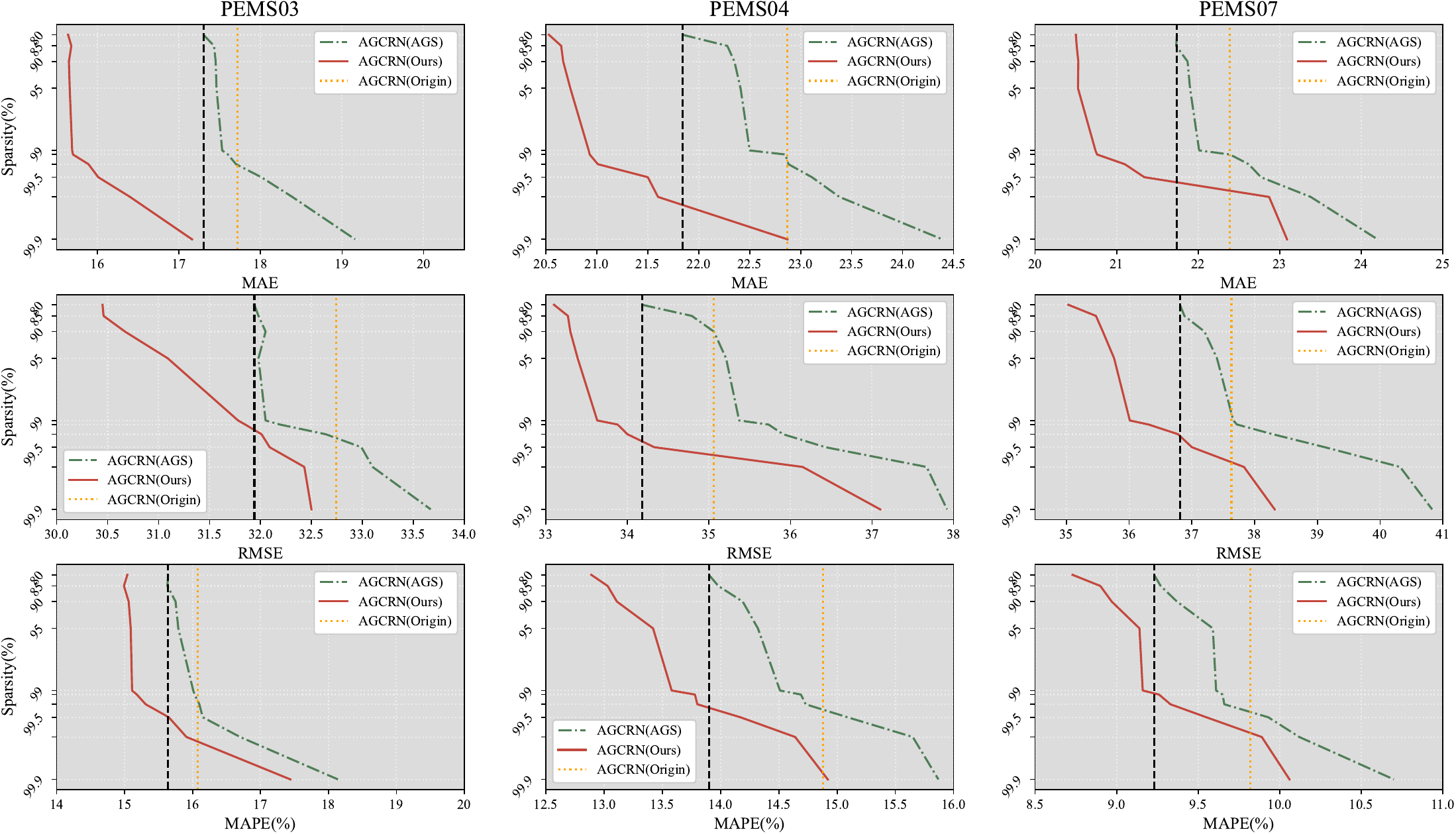}
    \caption{Test accuracies of STG-NCDE, localised by \textit{\sysname} and AGS, evaluated on transportation datasets. The black vertical dashed lines denote the accuracies of STG-NCDE when localised by AGS at a localisation degree of 80\%. The absence of STG-NCDE on the GLA dataset indicates that the model incurs out-of-memory issue. }
    \label{fig:traffic-stg}
\end{figure*}

Test performances in all metrics of AGCRN and STG-NCDE, localized by \textit{\sysname} and AGS across localisation degrees from $80\%$ to $99.9\%$, are respectively presented in \figref{fig:traffic-agcrn} and \figref{fig:traffic-stg} for transportation dataset.

\begin{figure*}[t]   
  \centering            
  \subfloat[AGCRN]
  {
      \label{fig:covid-agcrn-app}\includegraphics[width=0.45\textwidth]{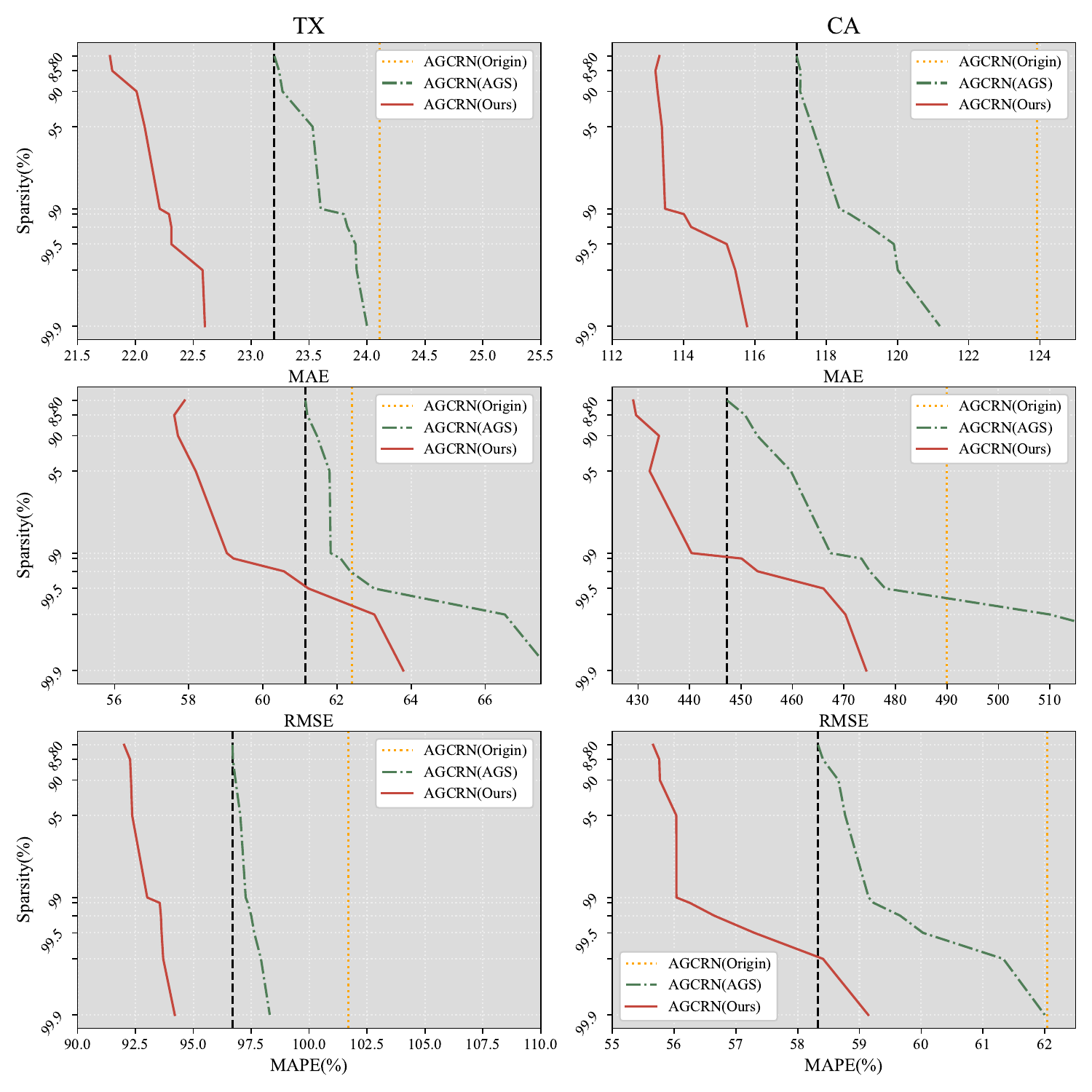}
  }
  \subfloat[STG-NCDE]
  {
      \label{fig:covid-stg-app}\includegraphics[width=0.45\textwidth]{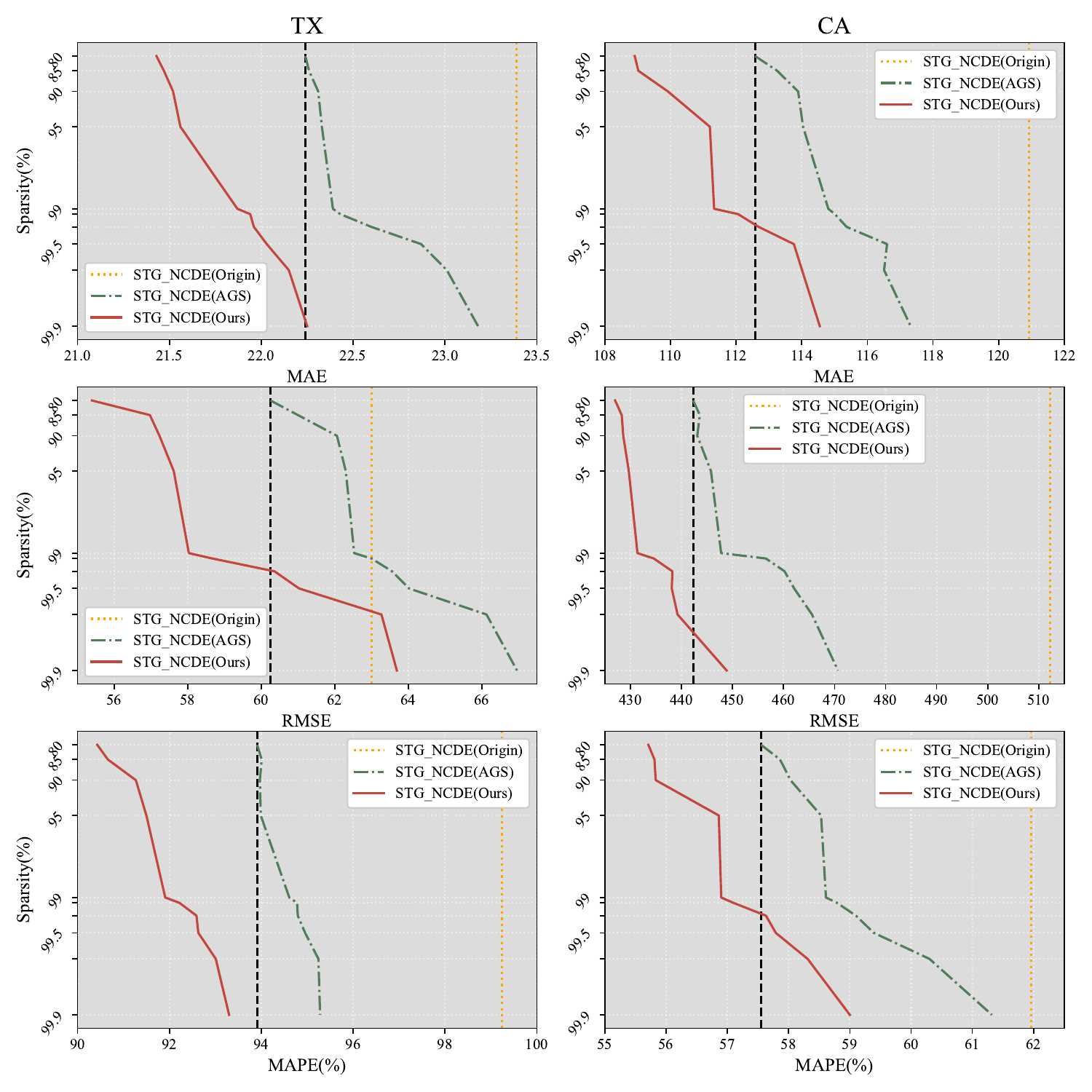}
  }
   \caption{Test accuracies of AGCRN (a) and STG-NCDE (b), localised by DynAGS and AGS, evaluated on biosurveillance datasets.The black vertical dashed lines denote the accuracies of AGCRN
 and STG-NCDE when localised by AGS at a localisation degree of 80\%.}    
  \label{fig:covid-app}   
\end{figure*}




\subsection{Description of Distributed Simulation}\label{app:sim}
We used Python’s Multiprocessing module to simulate distributed deployment, with each process representing a device. To facilitate and save memory, we created a shared memory ASTGNN model and dataset. It should be noted that in actual deployment, each device is equipped with an ASTGNN model and only has access to the data corresponding to its own nodes before sending communication requests. However, the simulation can still accurately calculate the communication overhead. 

     
      

\subsection{Personalised Localisation Experiment Details}\label{app: personal}

We categorised all nodes into four distinct groups, labeled as \Rmnum{1}, \Rmnum{2}, \Rmnum{3}, and \Rmnum{4}. In a personalised setup using \textit{\sysname}, these groups achieved in-degree compression ratios of 30\%, 50\%, 80\%, and 99\%, respectively.

To draw a meaningful comparison, for each group, we also trained a separate global-localised AGCRN using \textit{\sysname}, where every node in the network was subjected to the same in-degree compression ratio specific to that group. For instance, for Group \Rmnum{3} with an in-degree compression ratio of 80\%, we trained a model where all nodes, regardless of their group, had an in-degree compression ratio of 80\%.

The crux of our findings, depicted in \figref{fig:personal}, is that each group's performance in the personalised setting mirrored its performance in the corresponding global sparsity setup. For example, the Group \Rmnum{3} nodes in the personalised model performed as well as when every node in the system was globally set to an 80\% in-degree compression. This demonstrates that the individualised sparsity configurations of other groups did not hinder the performance of any specific group.

In essence, these results affirm that personalising the localisation of ASTGNNs doesn't compromise performance. Instead, it offers an effective and pragmatic approach.

\end{document}